\documentclass[a4paper,twoside]{article}

\usepackage{epsfig}
\usepackage{subcaption}
\usepackage{calc}
\usepackage{amssymb}
\usepackage{amstext}
\usepackage{amsmath}
\usepackage{amsthm}
\usepackage{multicol}
\usepackage{pslatex}
\usepackage{apalike}
\usepackage{algorithm2e}
\usepackage[bottom]{footmisc}
\usepackage{tipa} 
\usepackage[hyphens]{url}
\usepackage{multirow}
\usepackage{hyperref}
\usepackage{booktabs}
\usepackage{enumitem}

\usepackage{fancyvrb}
\usepackage{adjustbox}
\usepackage[utf8]{inputenc}
\usepackage{xcolor}
\usepackage{flushend}

\usepackage{listings}
\usepackage{caption}
\usepackage{float}

\lstdefinestyle{output}{
  basicstyle=\ttfamily\small,
  upquote=true,
  numbers=none,
  frame=none,
  breaklines=true,
  captionpos=b,
  aboveskip=0pt,
  belowskip=0pt,
  xleftmargin=0pt,      
  xrightmargin=0pt,
  columns=fixed,
  keepspaces=true
}

\lstdefinestyle{code}{
  basicstyle=\ttfamily\small,
  upquote=true,
  numbers=none,
  stepnumber=1,
  frame=single,
  breaklines=true,
  keywordstyle=\color{blue},
  commentstyle=\color{gray},
  stringstyle=\color{red}
}

\usepackage{SCITEPRESS}     

\newcommand{\gd}{G\`aidhlig}

\begin{document}

\title{A Rule-based Computational Model for \gd\ Morphology}

\author{\authorname{Peter J Barclay\sup{1}\orcidAuthor{0009-0002-7369-232X}}
\affiliation{\sup{1}School of Computing, Engineering, \& the Built Environment \\ Edinburgh Napier University, Scotland}
\email{email: p.barclay@napier.ac.uk}
}


\abstract{Language models and software tools are essential to support the continuing vitality of lesser-used languages; 
however, currently popular neural models require considerable data for training, which normally is not available
for such low-resource languages.
This paper describes work-in-progress to construct
a rule-based model of \gd\ morphology using data from Wiktionary, arguing that rule-based
systems effectively leverage limited sample data, support greater interpretability, and provide insights 
useful in the design of teaching materials. The use of SQL for querying the occurrence of different lexical patterns
is investigated, and a declarative rule-base is presented that allows Python utilities to derive inflected
forms of \gd\ words. This functionality could be used to support educational tools
that teach or explain language patterns, for example, or to support higher level tools such
as rule-based dependency parsers.
This approach adds value to the data already present in Wiktionary by adapting it to new use-cases.}

\onecolumn \maketitle \normalsize \setcounter{footnote}{0} \vfill

\section{\uppercase{Introduction}}
\label{sec:introduction}

\gd\ is a language used by a minority of people from Scotland. 
While online resources exist in the language, \gd\ shares a problem with other low-resource 
languages, in that there is a lack of data -- in this case, interpretable \gd\ text -- that can be used
to build modern tools such as grammar-checkers or dependency parsers.

This paper describes work-in-progress to leverage existing online \gd-language resources
by compiling information from Wiktionary into a standardised vocabulary format (SVF) that can be
used to model the morphology of the written language. The reformatted data was used in the following two ways.

First, the data facilitated creation of a
relational database of \gd\ lexemes. This database can be used for text analysis such as 
enumerating and quantifying various lexical forms, and comparing these to their occurrence in real 
\gd\ texts. Such analysis can help, for example, in the design of effective teaching materials
that emphasise common forms or draw attention to exceptions.

Secondly, the SVF data was used to inform Python utilities that can generate inflected forms
of \gd\ words using a rule-based notation. These utilities could support higher-level tools such as 
lemmatisers or rule-based dependency-parsers. They could also be used
to create interactive teaching tools that train or explain inflected words forms that 
appear in \gd\ texts.
This approach amplifies the value of the excellent
information and insights already contributed to \gd\ Wiktionary by its volunteer editors.

\section{\uppercase{Background}}

Celtic forms a sub-branch of the Indo-European language family; in classical times,
Celtic languages were spoken across a wide swathe of Europe, from modern
Ireland to modern Türkiye. The remaining Celtic languages all developed
in the British Isles\footnote{
Breton is not descended from the language of the Gauls, but from the Brittonic
languages of the south-west British Isles.}
and fall into two groups: Brythonic (Welsh, Cornish, and Breton) and Goidelic 
(Irish, \gd, and Manx). 

Many Indo-European languages share a common set of features, 
exemplifying as Standard Average European (SAE) typology \cite{haspelmath_european_2001}. 
However, Celtic and other ``fringe languages'' deviate from this pattern,
though to what degree is still a matter for debate \cite{irslinger_standard_2013}.
Moreover, the insular Celtic languages
share with unrelated Semitic languages such as Classical Arabic
a constellation of typological features including VSO word-order, inflected prepositions, and resumptive 
relative clauses -- an observation that has been 
discussed since the 17th century without any agreed explanation emerging.
A detailed discussion can be found in \cite{gensler_typological_1993}.
In any case, Celtic languages have some unusual linguistic features which makes
modelling their morphology an interesting problem. 

This work focuses on \gd, a Goidelic language spoken 
in Scotland (with some native speakers also in Canada). 
Customary usage in Scotland is to call the language `Gaelic',
but pronounced 
\textipa{/{\textquotesingle}\textg\textscripta l\textsci k/} 
when referring to the modern Scottish language, while pronouncing the same word
\textipa{/{\textquotesingle}\textg eIl\textsci k/} 
in the broader sense of the \gd, Irish and Manx languages together, or their associated literature and cultures. Since this difference is not apparent in writing, I the follow
recent academic convention of writing the name as `\gd'. While Irish, Manx and \gd\
are related languages, they are written differently, with
the divergent orthography of Manx especially obscuring its underlying
similarity.

Once spoken over most of Scotland and beyond, the 
\gd\ speech community has greatly declined, owing partly to normal
social and linguistic trends, especially with a language as important as 
English on the doorstep, and partly as a result of deliberate policies of institutional
suppression. Within living memory, children would be subjected to 
corporal punishment simply for speaking \gd\ in the classroom,
and the shadow of such persecution can be seen in 
lingering attitudes that may devalue
the language.

Nonetheless, the language is still in use, enriched by a long tradition of 
scholarship, oral and written literature, and music. 
Gaelic has the oldest literature in Northern Europe, pre-dated only by Greek and Latin in Europe as a whole. 
A good brief overview of the language and its history can be found in~\cite{mackinnon_lions_1974}.

Suppression has now largely been replaced by support, with organisations such
as An Comunn Gàidhealach\footnote{\url{https://www.ancomunn.co.uk/}}, 
BBC Alba\footnote{\url{https://www.bbc.co.uk/tv/bbcalba}}, 
and Comhairle nan Leabhraichean\footnote{\url{https://www.gaelicbooks.org/}} 
promoting the language. The 2022 Scottish census showed a modest increase in the number of people with 
some knowledge of 
\gd\footnote{\url{https://www.scotlandscensus.gov.uk/news-and-events/scotland-s-census-religion-ethnic-group-language-and-national-identity-results}}, though still only 2.5\% of the population. Duolingo's
website\footnote{\url{https://www.duolingo.com/courses}
(Dec 2025)}
shows over 600,000 people now studying \gd\ (though the number of active learners is surely smaller).
A number of schools now offer \gd-medium instruction, and \gd\ colleges have been established at
Sabhal Mòr Ostaig\footnote{\url{https://www.smo.uhi.ac.uk/}}
in Scotland at St.\@Anns\footnote{\url{https://gaeliccollege.edu/}} in Canada.
The language was finally given clear official status
in the Scottish Languages Act of 
2025\footnote{\url{https://www.legislation.gov.uk/asp/2025/10/part/1/chapter/1/crossheading/status-of-the-gaelic-language/enacted}}. 

With increasing interest in and appreciation of the language, the importance of appropriate learning materials
and software tools is growing. The work reported here is a small step towards their creation.

\section{About \gd}

\subsection{Written \gd}

\gd\ is written with an alphabet of 18 Roman letters, and uses grave accents to mark long vowels.
In Gaelic, as in Italian, the vowels O and E can have two pronunciations -- either a more open
or a more closed variant -- making the pronunciation of some words hard to determine unless the vowel quality is 
marked. Traditional \gd\ orthography marked this difference by using also an acute
accent, but this distinction was lost with the introduction of modern \gd\ orthographic conventions (GOC) in the spelling reform of 1981 \cite{sqa_gaelic_2009}.
A comprehensive overview of the use of accents in written \gd\ can be found
in \cite{ross2016standardisation}.

Here I focus on the written language, where accents can pose problems because
in some written texts we may find the same word accented according to the modern convention
or older conventions, or the accent may even be casually dropped. 
So for example, the word for big
may appear as \textit{mòr}, \textit{mór}, or \textit{mor}, depending on the age and provenance of the text.
Moreover, some words such as \textit{céilidh}
have been borrowed into English and are so well known
with the acute accent that they may be considered `frozen' in the earlier
orthography; we therefore need to consider them acceptable variants 
if they appear in \gd\ text.

\subsection{\gd\ Morphology}

\gd\ is a moderately inflected language, with two grammatical genders and 4 noun cases. One case, the vocative, is 
used in forms of address, and so mostly occurs in a few expressions referring (normally) to human beings.
Like most European languages, \gd\ distinguishes singular from plural nouns;
it also has a dual number, used when referring to exactly two things. 
However, its use is now largely limited to 
some stereotypical expressions referring
to things that usually come in twos (such as shoes); as it does not 
introduce new any lexical forms, 
the dual is not given further consideration here.

In common with many Indo-European languages, \gd\ words can take an suffix when inflected for different grammatical forms, such as \textit{saoghal + an} $\rightarrow$ \textit{saoghalan} (world $\rightarrow$ worlds). The last vowel or vowel group can be slenderised, such as \textit{fear} $\rightarrow$ \textit{fir} (man $\rightarrow$ men). 
In common with other Celtic languages, the beginnings of words also mutate by processes such as lenition (softening) (\textit{mo} + \textit{cat} $\rightarrow$  \textit{mo chat}), prothesis 
(\textit{ar} + \textit{iasg} $\rightarrow$ \textit{ar} \mbox{\textit{n-iasg}}), and glottalisation (\textit{òl} + past tense $\rightarrow$ \textit{dh'òl}).
Lenition is both a morphosyntactic and an allomorphic process: in the sentence \textit{thuit mo chat} (my cat fell), \textit{tuit} is lenited because it is a past tense, while \textit{cat} is lenited 
simply because it follows \textit{mo}.

\gd\ has a definite
article which appears in several forms (\textit{an, am, a', na, nan, nam}), but there is no indefinite article.
The presence or absence of an article 
or other particle can trigger lenition or 
prothesis in some combinations of words.

Table~\ref{tab:saoghal} shows an example noun inflection,
with singular and plural
forms of the word \textit{saoghal} (world) in different cases\footnote{Adapted from 
the Wiktionary table at \url{https://en.wiktionary.org/wiki/saoghal\#Scottish_Gaelic}.}.

\begin{table}
\centering
\begin{tabular}{|l |l |l|}\hline 
\multicolumn{3}{|l|}{\textbf{indefinite}} \\\hline
 & \textbf{singular} & \textbf{plural} \\\hline
nominative & saoghal & saoghalan \\\hline
genitive & saoghail & shaoghalan \\\hline
dative & saoghal & saoghalan \\ \hline \hline

\multicolumn{3}{|l|}{\textbf{definite}} \\\hline
 & \textbf{singular} & \textbf{plural} \\\hline
nominative & (an) saoghal & (na) saoghalan \\\hline
genitive & (an) t-saoghail & (nan) saoghalan \\\hline
dative & (an) t-saoghal & (na) saoghalan \\ \hline \hline

\multicolumn{3}{|l|}{\textbf{vocative}} \\\hline
 & \textbf{singular} & \textbf{plural} \\\hline
vocative & shaoghail & shaoghalan \\ \hline
\end{tabular}
\caption{Inflected forms of \textit{saoghal} (world).}
\label{tab:saoghal}
\end{table}

For text recognition,
prothesis can be removed during tokenisation, though of course it would
need to be correctly reproduced for text generation. For simplicity,
I concentrate here on the indefinite forms, and
for concision I will sometimes abbreviate the case names to two letters, for example  DP for `Dative Plural', and use GR for 'Gender'. 

Table~\ref{tab:saoghal} demonstrates that there is not a one-to-one
correspondence between grammatical forms and lexemes. There are 8 grammatical forms shown, represented by 5 lexemes, with another lexeme possible but not included in the table. This is summarised below:

\begin{itemize}[nosep]
    \item saoghal -- NS, DS
    \item saoghail -- GS
    \item saoghalan -- NP, DP
    \item shaoghalan -- GP, VP
    \item shaoghail -- VS
\end{itemize}
Text may also  contain \textit{shaoghal}, an allomorphic variation of the NS form, \textit{eg}.\@ \textit{mo shaoghal} (my world).

Many Indo-European languages exhibit a one-to-many relationship between lexemes and grammatical forms: the same written word may represent
more than one grammatical form. Owing to processes such as lenition
and prothesis however, in \gd\ this relationship is many-to-many: the same
written word may represent more than one grammatical form, and the 
same grammatical form may be represented by different written words.


Table~\ref{tab:ol} shows the inflection of the verb \textit{òl} (drink)\footnote{
Adapted from \url{https://en.wiktionary.org/wiki/\%C3\%B2l}.},
where 24 grammatical forms are represented by 16 lexemes plus one variant form. 
(The imperative singular is taken as the stem for other
verb forms, so `stem' is not counted as a separate form).
Although a \gd\ verb has more lexical variants than a 
noun, only the verbal noun is unpredictable. Apart from the case of a very few irregular verbs, the 
verbal forms are therefore easily derived.

\begin{table*}
\centering
\begin{tabular}{|l|l|c|c|c|}
\hline
\multicolumn{2}{|l|}{stem} & \multicolumn{3}{c|}{\textbf{òl}} \\ \hline
\multicolumn{2}{|l|}{verbal noun} & \multicolumn{3}{c|}{òl} \\ \hline
\multicolumn{2}{|l|}{past participle} & \multicolumn{3}{c|}{òlta} \\ \hline
\multicolumn{2}{|l|}{} & \multicolumn{2}{c|}{\textbf{independent}} & \textbf{dependent} \\ \hline
 &  & \textit{active} & \textit{passive} & \textit{active} \\ \hline
\multicolumn{2}{|l|}{past} & dh'òl & dh'òladh & dh'òl \\ \hline
\multicolumn{2}{|l|}{future} & òlaidh & òlar òltar & {òl} \\ \hline
\multirow{3}{*}{conditional} 
 & 1st singular & dh'òlainn & \multirow{3}{*}{dh'òltadh} & òlainn \\ \cline{2-2} \cline{3-3} \cline{5-5}
 & 1st plural & dh'òlamaid &  & òlamaid \\ \cline{2-3} \cline{5-5}
 & 2nd \& 3rd & dh'òladh &  & òladh \\ \hline
\multicolumn{2}{|l|}{relative future} & dh'òlas & dh'òlar &  \\ \hline
\multirow{6}{*}{imperative} 
 & 1st singular & òlam & \multirow{6}{*}{òlar òltar} &  \\ \cline{2-3}
 & 2nd singular & òl &  &  \\ \cline{2-3}
 & 3rd singular & òladh &  &  \\ \cline{2-3}
 & 1st plural & òlamaid &  &  \\ \cline{2-3}
 & 2nd plural & òlaibh &  &  \\ \cline{2-3}
 & 3rd plural & òladh &  &  \\ \hline
\end{tabular}
\caption{Conjugation of the verb \textit{òl}.}
\label{tab:ol}
\end{table*}

\gd\ distinguishes broad vowels (A, O, and U) from slender vowels (I and E), and the orthography adheres to vowel harmony -- within a word, wherever the pattern vowel-consonant-vowel occurs, the vowels must be both broad or both slender\footnote{
Leathan ri leathan is caol ri caol, leughar is sgriobhar gach facal 'san t-saoghal 
(Broad to broad, and slender to slender, (so) is read and is written every word in the world).}.
This means that there are alternative forms for all word endings to maintain vowel harmony on affixing, doubling the number of endings in use. 
For example, the common plural ending \textit{-an} (broad) also has the form \textit{-ean} (slender), and the verbal ending \textit{-adh} (broad) also has a slender form \textit{-eadh}.

The work reported here addresses morphology only; extending the approach to modelling the grammatical relationship between different lexemes awaits further investigation.
Discussing morphology, I use the word `lemma' to mean the dictionary form of a word, and
`lexeme' to  mean any form that may appear in a text, which may be the lemma or any of 
its inflected forms.
Apart from a few truly irregular forms, the inflection of most \gd\ words fall into a number of 
patterns; the difficulty is to know which pattern applies. Dictionaries often quote the 
`principal parts' for each word, which provide sufficient information to derive all the 
other forms. For nouns, the principal parts are the nominative singular (NS), 
the nominative plural (NP), and the genitive singular (GS). For verbs, 
the principal
parts are the singular imperative and the verbal noun (VN); and for adjectives, 
they are the positive and comparative (CP) forms. 

\subsection{\gd\ Wiktionary}
\label{sec:wikt}

The definitive \gd\ dictionary is `Dwelly', named after Edward Dwelly
who compiled it, and containing over 70,000 entries; it was first
published in 1920. Online \gd\ resources include \gd\ Wikipedia and Am Faclair Beag\footnote{\url{https://www.faclair.com/}}, 
an online dictionary which incorporates Dwelly, and the Digital Archive of Scottish Gaelic (DASG)\footnote{\url{https://dasg.ac.uk/en}}.
While Wikipedia and Wiktionary both
have permissive licences that allow re-use of their data, Wikipedia often contains
varied mixed-language text, and Wiktionary is more structured and focuses on words and definitions:
therefore it was selected as the source of the data used for this study.

Wiktionary\footnote{\url{https://www.wiktionary.org/}}
is an online, multilingual dictionary launched by the Wikimedia Foundation in 2002. It has entries in over 170 languages, including \gd, and is maintained by a community of volunteer editors. 
Comparing Wiktionary statistics, available at
\url{https://meta.wikimedia.org/wiki/Wiktionary/Table} (accessed Dec 2025),
highlights \gd's status as a `low-resource language'. While over 4 million editors
have created over 1 million pages for English, only 4000 users have contributed
around 12,000 pages for \gd, as shown in Table~\ref{tab:wikt}.
The number of active \gd\ contributors is considerably smaller still.

\begin{table}
\begin{adjustbox}{width=\columnwidth}
\begin{tabular}{|l |l |l |l |l|}\hline 
 & \textbf{entries} & \textbf{pages} & \textbf{edits} & \textbf{users} \\\hline
EN & 8646177 & 10501232 & 88596768 & 4349108 \\\hline
GD & 2950 & 12305 & 85993 & 4116 \\ \hline
\end{tabular}
\end{adjustbox}
\caption{Wiktionary stats -- English \textit{vs}.\@\gd.}
\label{tab:wikt}
\end{table}

The \gd\ entries cover many common words, showing inflections and giving usage examples.
The quality of the entries is generally high, and we can expect that they are well selected,
reflecting current usage and/or important words and expressions.
While this is an excellent \textit{ad hoc} reference for the language, it is hard gain a synoptic view of the data present. 

Compiling the \gd\ Wiktionary data into a structured format would add
value, and is the basis of the work described in the following sections.

\section{\uppercase{Aims  \& Approach}}
\label{sec:aims}

A rule-based approach to modelling language morphology has several
advantages, especially for low-resource languages. In addition to requiring less data than 
neural approaches, we note that:

\newpage
\begin{itemize}
    \item The model can provide a lower-level morphology stack for rule-based tools such as lemmatisers and parsers;
    \item since lexemes are defined by rules, the forms handled are inherently explainable;
    \item modest computational resources are required to run the model, making experimentation and deployment easier;
    \item the model can be used to support analysis of the appearance of inflected forms in \gd\ text, for example giving insight into how the language is evolving, by comparing time-stamped texts;
    \item moreover, such analysis would support design of effective teaching 
    materials and learning tools.
\end{itemize}

To paraphrase an old saying, ``a rule is worth a thousand data''. 
Combining the knowledge embedded in \gd\ Wiktionary with the rules of \gd\ 
morphology provides an approach to building natural language processing (NLP) utilities 
for use with \gd\ that complement neural models while requiring less data and offering
greater interpretability. 

While here I focus on \gd, the same approach may be beneficial for other typologically similar languages with limited text corpora.

This work aims to add value
to the body of knowledge compiled by the efforts of the volunteer
editors of \gd\ Wiktionary by adapting their data to create a rule-based 
model of \gd\ morphology.

\section{\uppercase{Methodology}}
\label{sec:method}

\begin{figure}
    \centerline{
    \includegraphics[width=\columnwidth]{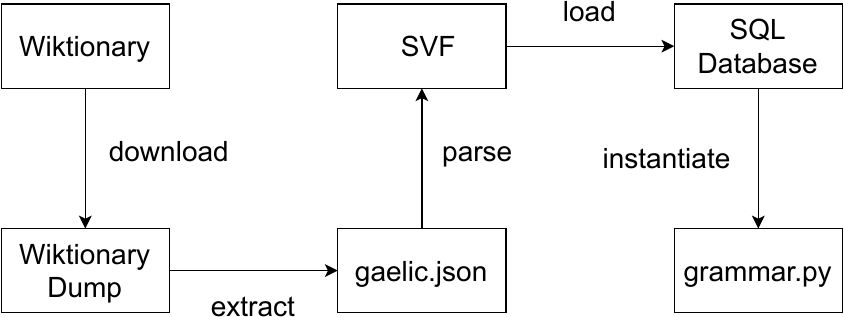}
    }
    \caption{Data preparation pipeline.}
    \label{fig:data-proc}
\end{figure}

The broad approach is to parse a dump of Wiktionary to filter out all \gd\ data, 
then parse these extracted entries to find the principal parts of each headword; these are then exported 
to a text file in a structured vocabulary format (SVF). This text file can then be used to 
create relational database tables, or 
populate a Python collection of objects representing
the principal parts of each word. The data processing pipeline is shown in
Figure~\ref{fig:data-proc}.

\subsection{Data Preparation}
\label{sec:prep}

I address the morphology of `content words' only, as these carry the
semantic weight of the text, while `function words' (or `stopwords') such as prepositions and conjunctions define
the grammatical relationships between these content words.
Therefore, only nouns, verbs, and adjectives were extracted from the Wiktionary dump.
For \gd, adverbs do not require special consideration
as they are regularly formed from adjectives using the particle \textit{gu};
for example. \textit{math} $\rightarrow$ \textit{gu math} (good $\rightarrow$ well).

The latest Wiktionary data dump was downloaded from \url{https://dumps.wikimedia.org/enwiktionary/latest} (accessed Dec 2025) for use
as the data source. The most up-to-date file is always called
\path{enwiktionary-latest-pages-articles-multistream.xml.bz2}; the instance
downloaded had an MD5 hash of  \texttt{ba88dcdfb6914bc487fa6a79ef1922b9} and the compressed size of the file was 1.7GiB.
Several open-source tools were
trialled for the data extraction, but these were not able to process
the file on a standard laptop, so a custom
Python script was used to extract all \gd\ entries, processing the file
in 200 MB chunks, and ensuring that data extracted from each chunk were dovetailed correctly in the output file.
Parsing the dump required several iterations, with frequent checking and 
amendment of the vibe-coded parser, as the format of the Wiktionary entries was not especially consistent.

Parsing a total of 10,247,958 XML  pages yielded 16,704 \gd\ entries.
The (uncompressed) Wiktionary dump of 11 GiB was distilled to a
JSON file of 17 MiB when only the \gd\ entries were extracted.

The JSON file derived was then parsed by a Python script to extract data for nouns, verbs,
and adjectives, writing their principal parts to a text file in a standardised format. 
Inflected forms that were separately listed in Wiktionary were subsumed
under their lemma, and explanatory notes and usage examples were skipped during this process.

The extracted vocabulary contained 
4956 nouns,  1025 adjectives, and
534 verbs, with 24 irregular forms noted.
Any principal parts that could not be sourced
from the Wiktionary entry were replaced by `?'.
The low number of verbs shown is perhaps unsurprising because,
while English often uses zero-suffix verb derivation
\cite{barbu2023semantic}, by contrast
\gd\ uses many light-verb expressions where a noun
carries most of the meaning;
for example, `I hope' is expressed as \textit{tha mi’n dòchas}, literally 
`I am in hope', where \textit{dòchas} (hope) is a regular noun and is recorded as such in the dataset.

The layout of the SVF text file created is
shown in Listing~\ref{lst:svf}.

\begin{samepage}
\begin{lstlisting}[caption={Layout of SVF.}, label={lst:svf},
frame=single, captionpos=b]
[Lemma] 
[POS] 
[Gender if NOUN] 
[Nominative Plural if NOUN] 
[Genitive Singular if NOUN] 
[Verbal Noun if VERB] 
[Comparative if ADJECTIVE]
[IRREG if irregular]
\end{lstlisting}
\end{samepage}

\begin{figure}
    \centerline{
    \fbox{\includegraphics[width=\columnwidth]{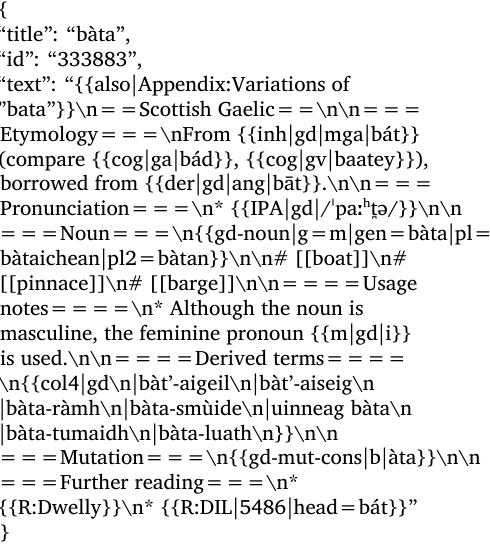}}
    }
    \caption{Wiktionary JSON entry for bàta (a boat).}
    \label{fig:bata-json}
\end{figure}

Taking for example the noun \textit{bàta} (a boat), the Wiktionary dump contains the complex and convoluted JSON shown in Figure~\ref{fig:bata-json}.

This was extracted and reformatted to produce the following SVF entry:
\begin{verbatim}
NOUN M "bàta" "bàtaichean" "bàta"
\end{verbatim}

When all the structured data had been compiled, it was
ready for cleaning and loading into a database.

\subsection{Data Completeness \& Data Cleaning}  

After extracting the relevant forms, 
a significant proportion of nouns and adjectives had at least one principal part absent,
marked by \texttt{"?"} in the SVF file. While this did not hinder the experiments presented in Section~\ref{sec:exp},
some implications and possible quality improvements are discussed in Section~\ref{sec:fw}.

The extracted SVF file also contained some near-duplicates that differed only by capitalisation 
(26 pairs) or by an accent (96 pairs). These numbers were sufficiently low to allow
manual data-cleaning.

Where entries differed only by capitalisation, some examples may have indicated
a specific nuance, for example \textit{Dia} (God -- a specific conception) versus
\textit{dia} (a god -- in general); some were lower-case variants of words that would normally
be capitalised, for example \textit{suaineach} for \textit{Suaineach} (Swedish); and others
seemed entirely to be random variants. All these pairs were reduced to a single 
preferred entry.

Where pairs differed only by an accent, this could indicate that they are different words, or
that one word is a recognised variant of the other.
On manual inspection, 68 pairs represented variants (the same word with a different spelling, including \textit{céilidh} mentioned above),
and 28 represented entirely different words. Where the pair represented different words,
both were of course valid entries; pending a more systematic treatment of variants,
these were also left in the dataset as separate lemmas, as they
are likely to be encountered in real texts.

\section{\uppercase{Experiments with the Data}}
\label{sec:exp}

With the data in a structured format, it was suitable for manipulation and experimentation. Section~\ref{sec:sql}
discusses analysing patterns with the data in relational format, and in Section~\ref{sec:py} I explore how the
data can be used with Python and a rule-based notation for modelling \gd\ inflections. 

\subsection{\gd\ Vocabulary with SQL}
\label{sec:sql}

The SVF notation can easily be modelled in a relational database and queried with SQL. 
Designing a database for principal parts presents one simple choice: 
whether to use one entity to represent all words, or one each for nouns, verbs, and adjectives. These two designs would be logically equivalent, and given the size of the dataset, 
performance is not likely to be an issue. 

Using three entity classes, as sketched in 
Listing~\ref{lst:3ent}, may seem natural, 
and this approach enforces semantic correctness (schema integrity) as 
each entity type can contain only the expected attributes.

However, it is probably more convenient to create a single entity class, as we will often process different types of words together.
The single table shown in Listing~\ref{lst:tb-def} was therefore created to hold the principal parts from each SVF entry.

\begin{lstlisting}[style=code, caption={Possible three-entity schema.}, label={lst:3ent},
frame=single, captionpos=b]
NOUN(Lemma, GR, NP, GS) 
ADJ(Lemma, CP)
VERB(Lemma, VN)
\end{lstlisting}

\begin{lstlisting}[style=code, language=SQL, caption={Table definition: Facal.}, label={lst:tb-def},captionpos=b, basicstyle=\ttfamily\small]
CREATE TABLE Facal(
    ID INT PRIMARY KEY AUTO_INCREMENT,
    Lemma VARCHAR(35) NOT NULL, 
    IRREG Bool DEFAULT FALSE,
    POS ENUM ('NOUN', -- part of speech
              'VERB', 
              'ADJ') NOT NULL,  
    GR ENUM ('M', 'F'), -- gender
    NP VARCHAR(35), -- nom. pl.
    GS VARCHAR(35), -- gen. sg.
    CP VARCHAR(35), -- comparative
    VN VARCHAR(35)) -- verbal noun
\end{lstlisting}

Although the SVF file contains only the correct principal 
parts according to each part of speech -- for example, verbal nouns are listed only where the POS is NOUN -- we can 
enforce semantic integrity for each 
part of speech with the following logical constraint expressed as a conjunction of three disjunctive normal forms,
one for each part of speech:
\begingroup
\small
\[
\begin{aligned}
\bigl(
  \bigl(
    &(\mathrm{GR} = \mathrm{NULL}
      \land \mathrm{NP} = \mathrm{NULL}
      \land \mathrm{GS} = \mathrm{NULL})
    {}\\
    &\qquad \lor \mathrm{POS} = \text{NOUN}
  \bigr)\\
  &\land (\mathrm{VN} = \mathrm{NULL}
    \lor \mathrm{POS} = \text{VERB})\\
  &\land (\mathrm{CP} = \mathrm{NULL}
    \lor \mathrm{POS} = \text{ADJ})
\bigl)
\end{aligned}
\]
\endgroup

This can be implemented as a CHECK constraint in SQL to ensure the schema 
of the corresponding database allows only
well formatted entries,
as shown in Listing~\ref{lst:tb-cc}.

\begin{lstlisting}[style=code, language=SQL, caption={SQL CHECK constraint for correct principal parts.}, label={lst:tb-cc},captionpos=b]
CHECK (
  ((GR IS NULL AND 
    NP IS NULL AND 
    GS IS NULL)
        OR POS = 'NOUN' ) AND
  (VN IS NULL OR POS = 'VERB') AND
  (CP IS NULL OR POS = 'ADJ')
  )    
\end{lstlisting}

The formatted data extracted from Wiktionary was then loaded into the database table that had been created.
With the data in relational format, we can analysis the frequency of different morphological patterns.

For example, a common way to form \gd\ plurals is by suffixing with \textit{-an} or a longer variant such as \mbox{\textit{-annan}} or \mbox{\textit{-aichean}}. 
By running a simple SQL query such as shown in Listing~\ref{lst:sql-pl-an}, we can establish that 2452 nouns
(of 4956 total, so over 49\%) follow this pattern; by changing $>=$ to $=$ we see that 1302 
of these use the shorter version of the suffix.

\begin{lstlisting}[style=code, language=SQL, caption={Counting nouns taking plural in \textit{-an}.}, label={lst:sql-pl-an},captionpos=b]
SELECT COUNT(NP)
FROM Facal
WHERE POS = 'NOUN'
AND LENGTH(NP) >= LENGTH(Lemma) + 2
AND NP LIKE '%an';
\end{lstlisting}

Taking another example, we can investigate the frequency distribution of different verbal noun patterns.
As noted earlier, these form are not predictable, but using \gd\ one notices that
\textit{-adh} is a very frequent ending for verbal nouns. We can quantify this observation with a simple SQL query such as
shown in Listing~\ref{lst:vn-end}.

\begin{lstlisting}[style=code,language=SQL, caption={Frequency distribution verbal noun endings.}, label={lst:vn-end},captionpos=b]
SELECT 
    SUBSTRING(VN, -3) AS Ending, 
    COUNT(SUBSTRING(VN, -3)) AS Freq
FROM Facal
WHERE POS = 'VERB'
AND LENGTH(VN) - LENGTH(Lemma) >= 3
GROUP BY Ending
ORDER BY Freq DESC;
\end{lstlisting}

The query is simplistic as it does not necessarily show the whole ending, and does not account
for forms where the stem might be modified on affixing.
Nonetheless, the result, shown in Listing~\ref{op:vn-end}, indicates that 218 verbal nouns of 534 (so around 40\%) follow this pattern,
with other endings falling into a handful of patterns and some unique forms occurring . 

Such observations can help, for example,
to design teaching materials that emphasise common patterns, group examples that follow the same pattern
for reinforcement, or identify unique patterns that require special attention.

\begin{lstlisting}[style=output, 
       caption={Verbal noun endings quantified by SQL.}, 
       xleftmargin=0.2755\columnwidth, 
       xrightmargin=\fill, 
       label={op:vn-end}]
+--------+------+
| Ending | Freq |
+--------+------+
| adh    |  218 |
| inn    |   18 |
| ail    |    7 |
| amh    |    6 |
| chd    |    4 |
| eam    |    2 |
| art    |    1 |
| ith    |    1 |
| eil    |    1 |
| idh    |    1 |
+--------+------+
\end{lstlisting}

The relational representation of the data could also be used for more detailed morphology analysis, 
or as a central store of principal parts that could be used by an interactive
\gd\ grammar website, for example, 
or by a Python morphology utility layer as described in
Section~\ref{sec:py}.

\subsection{Recognition-testing and Frequency Analysis}
\label{sec:rt-fa}

The model should now be able to recognise lexemes that appear in real \gd\ text. 
To test this, a database table \texttt{Occurrence }
was created from the frequency list published at
\url{https://github.com/innesmck/GaelicFrequencyLists},
derived from online \gd\ learning materials. This list was chosen as it is publicly available, and the lexemes are already tokenised and counted. The dataset contains 10,000 lexemes ranked by their frequency
in the source texts, representing 211245 total occurrences of these words.

Eyeballing the \texttt{Occurrence} table showed some duplicated or near duplicated entries,
and even some non-\gd\ words had slipped into the list.
However, for an initial experiment, I left these uncorrected 
as the majority of the 
entries seemed valid, and data drawn from real text will always be a little messy.
The table contained 690 hapax legomena; on inspection, these represented a mixture
or proper nouns, valid but less usual \gd\ words, and erroneous entries. 
These too were left uncorrected. 
The highest frequency lexemes
in the \texttt{Occurrence} table were all stopwords like \textit{e}, \textit{agus}, \textit{an}, and so on.

\begin{figure}[htbp]
\centering
\includegraphics[width=\columnwidth]{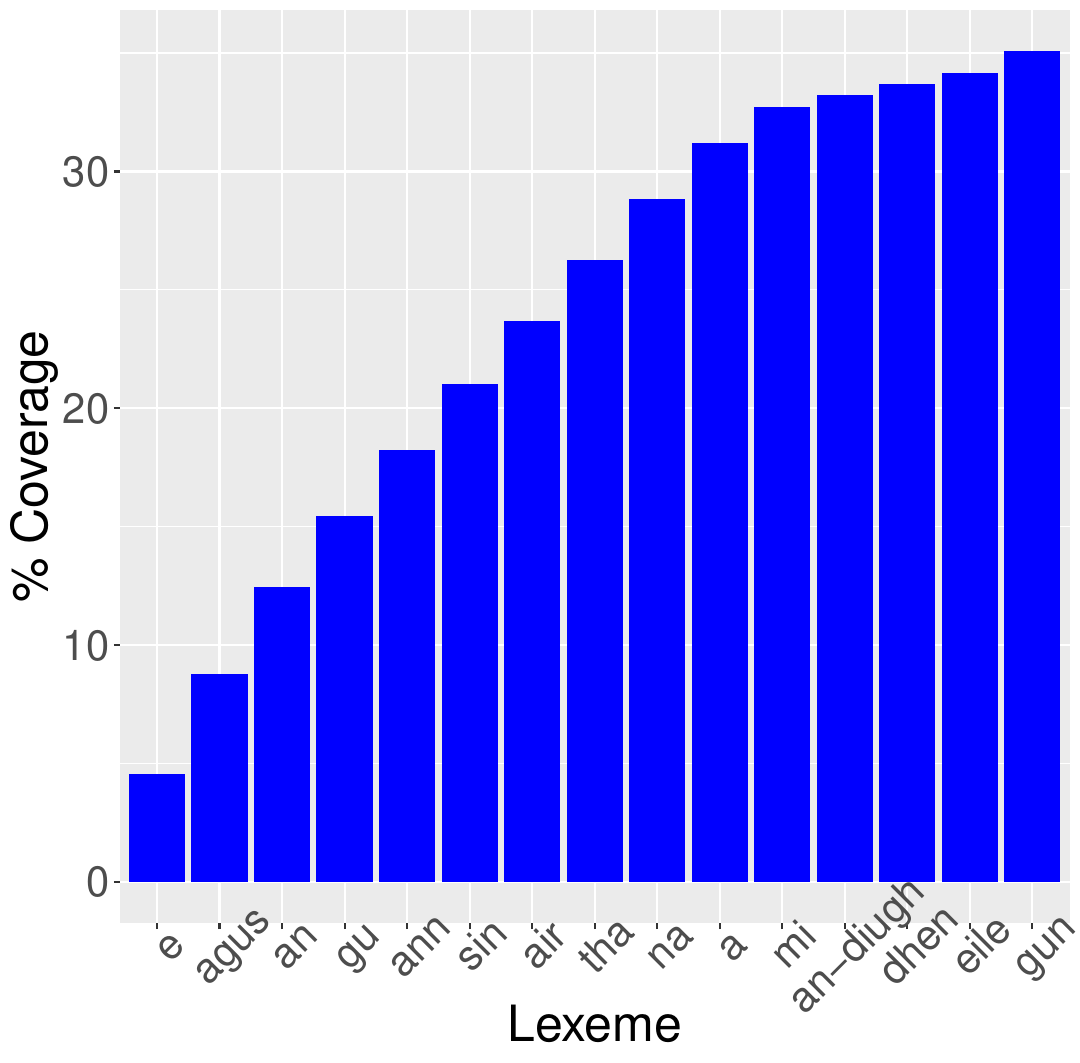}
\caption{Text coverage of first 15 stopwords.}
\label{fig:zipf}
\end{figure}

Zipf's law states that the frequency of a word in text is inversely proportional to its rank \cite{zipf_psychology_1946}. As a corollary, we expect that a 
significant amount of any text will usually be covered by a surprisingly 
small set of the most commonly occurring tokens. Of course, much of the meaning is carried by the 
long tail of less frequently occurring words.
Figure~\ref{fig:zipf} shows that the cumulative frequency of the first 15 
stopwords\footnote{Although \textit{tha} is really a verb, I left it in the list of stopwords as it is
used in so many constructions that, for purposes of frequency analysis, we could consider it
a grammaticalised particle.}
in the \texttt{Occurrence} table covers around 35\% of the text.
This sets the context for interpreting the level of coverage reported below
using the \texttt{Facal} table.

A simple SQL query was then used to establish how many of the lexemes
appearing in the frequency list could be matched to a lemma
from the standardised Wiktionary data in the \texttt{Facal} table, returning 2044
matches, so just over 20\%. Recalling that the \texttt{Facal} table contains only
content words from Wiktionary and no function words,
this level of matching is as expected. Many of the
commonest words in the original texts 
from which the \texttt{Occurrence} table was derived are of course stopwords such
as conjunctions and prepositions.

Next, a new table \texttt{AllForms} was created, which includes all
the lemmas from the \texttt{Facal} table, plus all their inflected forms
(without duplicates) which were derived and inserted by another 
Python script. This table contains all the distinct case
forms for nouns, and all the tense and mood forms for verbs.
For this purpose, all theoretical forms were
included, regardless of how likely they are to arise in
natural text. Comparing these two table, we see that 
33,132 unique inflected forms were generated from 6515 lemmas.

Now matching the entries in the \texttt{Occurrence} table against the \texttt{AllForms}
table achieved 4371 matches, so over 44\%. This demonstrates that the number of 
lexemes recognisable in the \texttt{Occurrence} table is more than 
doubled when matching against inflected forms, with around
half of the content words appearing in the original text
in lemma form. 

This quick experiment shows only that recognising inflected forms greatly expands the 
ability of the core dataset to recognise lexemes drawn from real texts.
Further analysis can determine which inflections appear most frequently and which are rare. 
Such insight could help structure learning
materiels, for example by presenting early for study the grammatical forms
that most increase the recognition rate of lexemes appearing
in real \gd\ texts.

\subsection{\gd\ Vocabulary with Python}
\label{sec:py}

The data in SVF notation can be used to instantiate Python objects representing different lexemes 
that may appear in \gd\ text.
A class \texttt{Facal} (word) was created to represent each word, along with a class \texttt{Faclair} (vocabulary) to hold a searchable collection of \texttt{Facal} instances.

After looking up a word object, its principal parts may be accessed; for example, where \texttt{f} is an object of class \texttt{Faclair}, the expression \texttt{f.lookup(\textquotesingle saoghal\textquotesingle ).NP} returns the nominative plural of this word, to wit \texttt{\textquotesingle saoghalan\textquotesingle }. 

Going further, we can now add methods to derive the non-principal parts according to known grammar rules. 
The class \texttt{Faclair} was therefore extended to read a knowledge-base of inflected forms, expressed in a simple rule-based notation, 
defining the other grammatical forms in terms of the principal parts. An example rule is shown 
in Listing~\ref{lst:rule}.

\begin{lstlisting}[style=code, 
       captionpos=b,
       caption={Rule to decline feminine nouns.}, 
       xleftmargin=0.0\linewidth, 
       xrightmargin=\fill, 
       label={lst:rule}]
* NOUN & F
NS: NS; NP: NP; GS: GS; GP: H/NP
DS: NS; DP: NP; VS: H/GS; VP: H/NP
\end{lstlisting}

The line with an asterisk shows which words will be matched; in this case, feminine nouns. 
In the following rules, 
the form on the left of : is the required form, the form on the right of : is a principal part.
Here, the \texttt{h/} indicates that the form should be lenited. Ordering of rules can be used
to ensure the correct rule is applied, with more specific rules applied first.
(A separate module, \texttt{Funcs.py}, defines functions for transformations such as lenition that can be applied
when generating the inflected forms).

So the example rule shown indicates that in this case the dative singular (DS) is the same as the nominative singular (NS),
or that the vocative singular (VS) is derived by leniting the genitive singular.
Inflected forms can be retrieved by the \texttt{inflect} method, so for example the expression 
\texttt{f.lookup(\textquotesingle saoghal\textquotesingle ).inflect(\textquotesingle DP\textquotesingle )} will return \texttt{\textquotesingle saoghalan\textquotesingle }.

With these forms derivable from the rule-base, a method \texttt{decline} 
was added to enumerate all forms of a given lemma.
This functionality can be demonstrated 
by the simple test program as shown in Listing~\ref{lst:decl}.

\begin{lstlisting}[style=code,
    language=Python, caption={Noun declination method.}, label={lst:decl},captionpos=b]
from Grammar import Faclair
f = Faclair()
word = 'saoghal'
f.decline(word)
\end{lstlisting}

This produces the tabular output shown in Listing~\ref{op:rb-dec}. \\

\begin{lstlisting}[style=output, 
       caption={Rule-based declination of \textit{saoghal}.}, 
       xleftmargin=0.1\linewidth, 
       xrightmargin=\fill, 
       label={op:rb-dec}]
|case | singular  | plural     |
+------------------------------+
| nom.| saoghal   | saoghalan  |
| gen.| saoghail  | shaoghalan |
| dat.| saoghal   | saoghalan  |
| voc.| shaoghail | shaoghalan |
+------------------------------+
\end{lstlisting}

\section{\uppercase{Overview of some Related Work}}
\label{sec:rw}

\subsection{NLP for low-resource Languages}

Transfer learning and a variety of other neural approaches have been applied to low-resource languages. 
Nehme \textit{et al.} report success in fine-tuning a model pre-trained on 
unrelated Romance languages to translate Akkadian \cite{nehme_translating_2025},
an extremely low-resource language. An effective method of applying transfer learning to various
low-resource languages, including \gd, is described in \cite{minixhofer_wechsel_2022}.
A good survey of the application of neural methods to low-resource languages in general can
be found in \cite{hedderich_survey_2021}.

However, the quality of neural models is dependent on the volume of data available
for training \cite{akhmetov_highly_2020}.  
Moreover, neural models gloss over linguistic fundamentals, and Meroni argues 
that the approach encourages ``a pursuit of superficial aesthetic results, neglecting the foundational linguistic structures that underpin natural language processing''
\cite{meroni_bringing_2025}. A comparison of neural an rule-based approaches can be found in
\cite{strubbe_rule-based_2024}.

\subsection{NLP for \gd}

A Scottish Gaelic Linguistic Toolkit\footnote{\url{https://www.vdu.lt/cris/entities/product/19691cad-b901-4f10-b65b-2bd1e57cdbd9}.
}, trained on the Annotated Reference Corpus of Scottish Gaelic (ARCOSG)\footnote{
\url{https://www.research.ed.ac.uk/en/datasets/annotated-reference-corpus-of-scottish-gaelic-arcosg}.}, 
contains utilities for tagging, lemmatisation and parsing \gd\ text.
Batchelor presented a treebank of universal dependencies for \gd\ \cite{batchelor_universal_2019},
and the UD approach has been used to create an online linguistic analyser \cite{utka_online_2020}.

While individual \gd\ utilities have been evaluated \cite{lamb_evaluating_2016},
there is a lack of literature comparing different approaches, though one recent paper did compare different POS taggers for Welsh \cite{prys_evaluation_2022}.
More systematic evaluations would be useful, especially as some taggers do not perform as well as publicised when presented with 
out-of-sample test datasets \cite{wahde_challenging_2024}.

Progress in NLP for low-resource languages rests on new methods and new data. 
While recent work proposes interesting approaches to
collecting useful \gd\ datasets, such as summarising 
live \gd\ dialogues \cite{howcroft_building_2023},
existing resources such as Wiktionary have not yet been fully exploited.

\section{\uppercase{Ongoing \& Future Work}}
\label{sec:fw}

This paper describes work in progress, and evaluation of the approach is still on-going.

Currently the rule-based system excludes wholly irregular 
forms -- fortunately these are few -- but they will need to be added as special cases. This will require a call
to special rules where the IRREG tag appears. Also, as noted, some words can appear in variant forms, and
these need to be handled more systematically, 
which may require a more nuanced database schema.
The rule-base still needs to be expanded to cover all common patterns of morphology, any edge-cases identified,
and a thorough assessment of its accuracy undertaken. This is work in progress.

One disadvantage of a rule-based approach is that we are very reliant on the quality of the data;
neural models trained with large data volumes can tolerate some degree of inaccuracy in their input.
When preparing the SVF file, some principal parts were not found in Wiktionary.
While all data was present for verbs, around 30\% of nouns had at least one principal part not found.
This could mean (1) the entry
is a compound word, and the principal parts are listed under its components; (2)
the form  is regularly derived, so not listed; (3) the form does not exist; or (4) 
the form does exist but is  simply missing.
Examples of forms that do not exist would include words such 
as \textit{airgead} and \textit{uisge} (money and water) which have no plural.
The quality of the data could be increased by marking such cases with \texttt{"-"} instead of \texttt{"?"} 
to show that the form is non-existent rather than absent.

One approach to filling in any genuinely missing forms would be to check for them in other reference
sources, whether manually or automatically, and starting with those with highest occurrence in real text.
Another and interesting approach could be to generate a missing form as if it were regular, then search for it 
in existing corpora; if it is found, especially in the expected context, it is likely correct. 
For example, in Wiktionary no comparative form is listed 
for \textit{fadalach} (late), but assuming it to be regular this would be \textit{nas fadalaiche} (later), so
we can search to see if this appears in any corpora (which it does). 
Following the particle \textit{nas}, we can be confident this is a comparative
form and add it to the database.

The approach of compiling structured data from semi-structured online resources has other  potential
applications, where data designed for \textit{ad hoc} reference can 
be repurposed for teaching and learning.
During this project, for example, I extracted the illustrative sentences from the
Wiktionary dump, and removing duplicates and ordering them by length, created a 
deck with 1080 phrases for the ANKI spaced-repetition software application.
This can be used for structured individual 
study of \gd\ phrases. 

Rule-based models could also be adapted to generate many varied examples of  
grammatically correct sentences which could then be used for data augmentation
when training neural models, an approach which has already shown success
with other low-resource languages 
\cite{lucas_grammar-based_2024}. Such generated sentences could also 
inform interactive teaching and learning materials.

Looking ahead, several stepping stones could advance the rule-based approach.
While neural models ingest continuous text, which may incur copyright issues,
the rule-based approach can be facilitated by 
corpora simply consisting of lists of lexemes and frequencies -- that is, ranked word lists.
However, for greater utility, it would be helpful to tag occurrences by provenance and date,
so statistics could be drawn from different subsets of the data as required.
Another useful resource would be a modern library of interoperable pipeline components that could
be composed and re-used in support of experimentation.
The recently published spaCy-compatible Scottish Gaelic NLP pipeline is a 
good step in this direction \cite{sgrp-core-web-sm_nodate}.

\section{\uppercase{Resources}}
\label{sec:res}

The AI-enhanced 
integrated development environment Cursor (\url{https://cursor.com/} was used to create and test code for parsing the Wiktionary dump. 
The database schema and SVF data files are available
on Zenodo at \url{https://zenodo.org/records/18319154} (DOI: 10.5281/zenodo.18319154).
The `shared deck' \textit{Gaelic Phrases from Wiktionary} for ANKI 
has been published on AnkiWeb 
at \url{https://ankiweb.net/shared/info/117198512}.  

\section{\uppercase{Conclusion}}
\label{sec:conclusion}

The definitions in Wiktionary are designed as an \textit{ad hoc} reference source.
By 
extracting this information and 
compiling
it in a structured format, this work seeks to add value 
by adapting the existing data to new use cases.
Early results show promise that rule-based models can
offer an approach
complementary to neural models,
supporting interpretable recognition and analysis of \gd\ morphology
despite limited resources, 
and
paving a path towards developing software tools and 
learning resources that can support the \gd\ language
into the next century.


\bibliographystyle{apalike}
{\small
\bibliography{GaelicRefs}}

\end{document}